\documentclass[11pt]{article}

\usepackage{bbold,bm,amsmath,amssymb,latexsym}
\usepackage{textcomp}
\usepackage[colorlinks = true, citecolor = blue]{hyperref}
\usepackage{bookmark}
\usepackage[numbers]{natbib}
\usepackage[all]{hypcap}
\usepackage{booktabs}
\usepackage{cite}{\tiny }
\usepackage{afterpage}
\usepackage{graphicx}
\usepackage{mdframed}
\usepackage[normalem]{ulem}
\usepackage[table]{xcolor}
\usepackage{makeidx}
\usepackage{cleveref}
\usepackage[centerlast]{caption}
\usepackage{float}
\usepackage[lined,boxed,commentsnumbered]{algorithm2e}
\usepackage[all]{xy}
\usepackage{tikz}
\usepackage{pgfplots}
\usetikzlibrary{arrows,intersections}
\usepackage{times}
\usepackage{float}
\usepackage{subfig}
\usepackage{authblk}
\usepackage{tikz-cd}
\usepackage[utf8]{inputenc}
\usepackage{}
\usepackage[T1]{fontenc}

\newlength\aftertitskip     \newlength\beforetitskip
\newlength\interauthorskip  \newlength\aftermaketitskip

\newcommand{\BlackBox}{\rule{1.5ex}{1.5ex}}  
\newenvironment{proof}{\par\noindent{\bf Proof\ }}{\hfill\BlackBox\\[2mm]}
 
\newtheorem{theorem}{Theorem}
\newtheorem{lemma}[theorem]{Lemma} 
\newtheorem{proposition}[theorem]{Proposition} 

\newtheorem{corollary}[theorem]{Corollary}
\newtheorem{definition}[theorem]{Definition}

\newcommand{\RR}{\protect\mathbb{R}}

\newcommand{\EE}{\protect\mathbb{E}}
\newcommand{\PP}{\protect\mathbb{P}}
\newcommand{\TT}{\protect\mathbb{T}}
\newcommand{\MM}{\protect\mathbb{M}}
\newcommand{\obs}{\protect\mathcal{Z}}

\newcommand{\ind}{\protect\bm{1}}

\newcommand{\dist}{\raise.17ex\hbox{$\scriptstyle\mathtt{\sim}$}}
\newcommand{\Lbar}{\protect{\underline{L}}}
\newcommand{\risk}{\protect{\mathrm{Risk}}}
\newcommand{\Rbar}[1]{\underline{\risk}_{#1}}

\newcommand{\norm}[1]{\left\lVert #1 \right\rVert}

\newcommand{\ip}[2]{\left< #1, #2\right>}

\newcommand{\dual}[1]{(\RR^{#1})^*}

\newcommand{\riskL}[1]{\risk_{#1}}
\newcommand{\pred}[1]{[\![ #1 ]\!]}

\newcommand{\id}[1]{\mathrm{id}_{#1}}

\pgfplotsset{compat=1.18}

\title{Beyond Blackwell: Constrained Data Processing Inequalities}

\author{Brendan van Rooyen\\
  Epigraph Solutions \\
  \texttt{brendan.vanrooyen@outlook.com}}
\date{}

\begin{document}
	
\maketitle

\begin{abstract}
We study the comparison of statistical experiments from a decision-theoretic standpoint, working with Markov transitions as basis-free linear maps on dual spaces, following \citep{JMLR:v18:16-315}. Our starting observation is that the data-processing inequality — that adding noise to an experiment cannot lower its Bayes risk, is the functoriality of fibrewise minimisation over a presheaf of decision rules: the rules available after an experiment form a presheaf, garbling acts on it by precomposition, and a risk functional is a natural transformation from it to the ordered reals. The argument refers to neither probabilities nor expected values, and it survives replacing the decision presheaf by a sub-presheaf of admissible rules. Any constraint imposed on the strategy that a rule induces — fairness, unbiasedness, is automatically of this form, and so yields a data-processing inequality with no counterpart in the unconstrained Blackwell theory.
\\
\\
Our main result is their quantitative companion. For constraints closed under post-processing we define a constrained deficiency and prove a constrained randomization theorem: one experiment dominates another on every admissible decision problem exactly when the constrained deficiency is small. The classical theorem is the case of no constraint, where the identity is itself an admissible rule; in general it is not, and the constrained deficiency is an optimisation over the coarsenings of an experiment that render it compliant. Specialising to equalized odds, we exhibit a pair of experiments that unconstrained Bayes risk identifies at a given loss and prior but that the constrained risk separates, giving a one-line certificate of Blackwell-incomparability. We work throughout with finite sets.

\end{abstract}

\section{Introduction}
Experiments link unknown states of the world to observable data. A basic question, going back to Blackwell \citep{Blackwell:1954}, is when one experiment is more informative than another regardless of the decision problem at hand. Blackwell's answer is that one experiment divides another when the second can be obtained from the first by adding noise; the operational content of this order is the data-processing inequality, that garbling an experiment cannot decrease the Bayes risk available to a decision maker.
\\
\\
The usual route to that inequality runs through decision rules, risk, and loss functions, and the conclusion can appear to depend on this machinery. It does not. The inequality is a structural fact about minimisation: the decision rules available after an experiment form a presheaf over the category of experiments under a fixed base of unknowns, garbling acts on this presheaf by precomposition, and a risk functional is a natural transformation from it to the ordered reals. The data-processing inequality is the functoriality of the induced fibrewise minimum, a statement mentioning neither probability distributions nor expectations. Classical inequalities for total variation and for f-divergences are instances.
\\
\\
The reason to say it this way is that the argument never used the fact that all decision rules were available. It used only that the available rules are stable under precomposition with a garbling, and this is a property that many practically motivated restrictions also have. Fix a set of admissible strategies, maps from unknowns to actions, those that are unbiased for some quantity, or fair with respect to a protected attribute. A decision rule is admissible after an experiment when the strategy it induces is admissible. Because precomposition by a garbling leaves the induced strategy unchanged, admissibility in this sense is automatically stable, so admissible rules form a sub-presheaf and inherit a data-processing inequality. Restricting a decision maker to fair rules does not cost them the guarantee that noise degrades performance.
\\
\\
Whether this is more than bookkeeping is a fair question.. Take a task label and a protected attribute, and compare an experiment that reports the label through a group-blind binary symmetric channel against one that reports it perfectly on one group and twice as noisily on the other. The two inject the same average noise, and their posteriors over the task label coincide, so their unconstrained Bayes risks agree at the natural loss and prior; the classical theory is silent. Under equalized odds, however, the admissible rules over the second experiment collapse to those ignoring the data entirely, and its constrained risk jumps to that of the trivial experiment. One constrained functional, evaluated at a single loss and prior, thus certifies that the two experiments are Blackwell-incomparable, where the Blackwell–Sherman–Stein theorem would otherwise require hunting for a decision problem on which the risks reverse.
\\
\\
The main contribution is the quantitative counterpart. Le Cam's deficiency measures how far one experiment is from being a garbling of another, and converts into a bound on the difference of Bayes risks; the randomization theorem says the two are equivalent. Both presuppose access to every decision rule. We define a constrained deficiency for constraints closed under post-processing and prove that it plays the same role, with a constrained Blackwell–Sherman–Stein theorem as the case of exact domination. The construction is an optimisation over the coarsenings of an experiment that render it compliant, and it suffices to range over the minimal such coarsenings. When the constraint is vacuous the identity is itself compliant and the classical deficiency is recovered.
\\
\\
Related work falls into two groups. The decision-theoretic and transition apparatus is not new: the dual-space view of probability and the algebra of transitions follow \citep{JMLR:v18:16-315}, the loss-and-risk setup follows \citep{Blackwell:1954,DeGroot1962,Le1964,Torgersen1991,Williamson2014}, and our transitions are Markov kernels presented without a fixed basis, equivalently the Markov category of stochastic relations; the closest categorical relative is the synthetic treatment of experiment comparison due to \citep{Fritz_2023}. Separately, several restricted variants of the Blackwell order are known, the Lehmann order, the strong Blackwell order, orders arising from elicitation, but these restrict the class of decision problems against which experiments are compared. We instead restrict the class of decision rules the decision maker may use, holding the decision problems fixed. The two axes are independent, and it is the second that produces sub-presheaves.

\section{Background}
We phrase the general decision problem as a two-player game between nature and a decision maker. Let $\Theta$ be a set of possible values of some unknown, and $A$ the set of actions available to the decision maker. The consequence of an action is measured by a loss function $L : \Theta \times A \rightarrow \RR$. A negative loss represents a gain to the decision maker. The norm of a loss function is given by its largest possible consequence (positive or negative), $\norm{L}_\infty = \max_{\theta, a} \lvert L(\theta,a) \rvert$.

\subsection{Probability}

The loss function itself does not tell us how to act; it provides a \emph{loss profile}, a function $L_a : \Theta \rightarrow \RR$, for each action $a \in A$. This loss profile summarizes the loss incurred for each unknown value for a fixed action. Our knowledge of $\theta$ allows us to place a value on each of these actions, through an \emph{expectation}. Here we place the restriction that the expectation should be given by a \emph{probability distribution}.

\begin{definition}\label{Probability Distribution}
	
	A \emph{probability distribution} on a set $X$ is an element of $\dual{X}$, i.e. a linear function $\EE_P : \RR^{X} \rightarrow \RR$, such that:
	
	\begin{enumerate}
		
		\item $\EE_P \ind_X = 1$.
		\item If $f(x) \leq g(x) ,\ \forall x \in X$ then $\EE_P f \leq \EE_P g$.
		
	\end{enumerate}
	
\end{definition}
Point one states that probability distributions are \emph{normalized}, point two that they are \emph{positive}. At times we use infix notation, with $\EE_{P} f = \ip{P}{f}$. Define the set of all distributions on a set $X$ to be $\PP(X)$. For any $x \in X$ define the point mass distribution $\delta_x$, with $\EE_{\delta_x} f = f(x)$ for all functions $f$.

\subsection{Experiments}
We have assumed that the decision maker represents their uncertainty in $\Theta$ through a probability distribution $P$. The key question is \emph{which} $P$. To determine $P$, the decision maker is guided by \emph{experiments}. In its simplest form, an experiment is a function $e : \Theta \rightarrow \PP(\obs)$, which assigns to each value $\theta$ a probability distribution on the results of the experiment. Functions of this type are called \emph{Markov Kernels}, and they can be understood by their action on functions and by extension their action on duals. Define,
\[
e^*(f)(\theta) = \EE_{z\dist e(\theta)} f(z).
\]
the pull-back of by $e$ of $f \in \RR^\obs$. Dually define the \emph{push-forward} by $e$, $e_*: \dual{\Theta} \rightarrow \dual{\obs}$ with,
\[
\ip{e_*(\mu)}{f} := \ip{\mu}{e^*(f)}.
\]
Rather than define a composition of Markov kernels directly as in other treatments, we rather equate the kernel with its push-forward map and use ordinary function composition. The usual composition rules then fall out as linear algebra subordinate to a particular basis.

\subsection{Transitions}

\begin{definition}
	A \emph{transition} from a set $X$ to a set $Y$ is a linear map $T : \dual{X} \rightarrow \dual{Y}.$
\end{definition}
Although abstract in appearance, we observe that when $X$ and $Y$ are \emph{finite}, a transition is nothing more than a \emph{matrix}. In general, a transition is an integral operator. Denote the set of all the transitions from $X$ to $Y$ by $\TT(X,Y)$. We call a transition \emph{Markov} if $T(\PP(X)) \subseteq \PP(Y)$. Denote the set of all Markov transitions from $X$ to $Y$ by $\MM(X,Y)$. When $X$ and $Y$ are finite, Markov transitions are represented by column stochastic matrices. The distribution,
$$
T(x) := T(\delta_x)
$$ 
is how the decision maker summarizes their uncertainty about $Y$ if the true value of $X$ is $x$.
For transitions $f \in \TT(X,Y)$ and $g \in \TT(Y,Z)$ we can define $g\circ f\in \TT(X,Z)$ by the composition of functions. 
\\
\\
The above can be understood as defining a particular category, with objects as finite sets and arrows as transitions. Restricting to Markov transitions yields a subcategory, sometimes called the category of stochastic relations. This category will be of our primary interest.

\subsection{Comparing Experiments}\label{sec:comparing-experiments}
An \emph{experiment} is a Markov transition $e \in \MM(\Theta, \obs)$. We call $\obs$ the observation space of the experiment. We assume that $\obs$ is finite. The distribution $e(\theta)$ summarizes the uncertainty of the decision maker in the observation when $\theta$ is the unknown value. Leaving $\obs$ to vary yields a category $\MM(\Theta,-)$, of elements under $\Theta$. Objects of this category are experiments with arrows given by commutative triangles,	
	\[
	\xymatrix{
		& \Theta \ar[ld]_e \ar[rd]^{e'} & \\
		\obs\ar[rr]^f && \obs'
	}
	\]
The Markov transition $f$ can be understood as adding noise to the experiment $e$, which results in the corrupted experiment $e'$. After observing the results of an experiment, the decision maker is tasked with choosing a suitable action. They do this through \emph{decision rules}.
\\
\\
A \emph{decision rule} is a Markov transition $d \in \MM(\obs, A)$. The distribution $d(z)$ summarizes the uncertainty of the decision maker in which action to choose, given an observation $z \in \obs$. We define the \emph{risk},
$$
\riskL{L}(\theta, e, d) = \EE_{a \dist d\circ e(\theta)} L(\theta,a).
$$
The risk measures the quality of the final action chosen by the decision maker when using the decision rule $d$, after performing the experiment $e$, assuming $\theta$ is the true value of the unknown. The risk does not provide a single number for the comparison of experiments; rather, it provides an entire \emph{risk profile}. To compare risks directly, the decision maker can use \emph{Bayesian} risk defined as,
\begin{align*} 
\riskL{L}^{\pi}(e,d) := \EE_{\theta \dist \pi} \riskL{L}(\theta,e,d). 
\end{align*}
This quantity allows the decision maker to \emph{compare} the usefulness of experiment, decision rule pairs. To compare experiments directly, we assume that the decision maker uses the \emph{best} decision rule. Define the minimum Bayesian risk as,
\begin{align*} 
\Rbar{L}^{\pi}(e) := \min_{d} \riskL{L}^{\pi}(e,d).
\end{align*}
The Bayes risk is more than a function, it is a \emph{functor}. Consider the category $(\RR, \leq)$, with objects real numbers and $f : x \rightarrow y$ if and only if $x \leq y$. Suppose $f : e \rightarrow e'$, then,
\[
\riskL{L}^{\pi}(e',d')  = \riskL{L}^{\pi}(e,d'\circ f).
\]
Taking minima yields,
\[
\Rbar{L}^{\pi}(e') = \min_{d'}\riskL{L}^{\pi}(e,d'\circ f) \geq \min_{d}\riskL{L}^{\pi}(e,d) = \Rbar{L}^\pi(e).
\]
Therefore $\Rbar{L}^\pi$ is a functor from $\MM(\Theta, -) \rightarrow (\RR, \leq)$. This is sometimes called the \emph{data processing theorem} or the \emph{data processing inequality}.

\subsection{Most and Least Informative Experiments}

For any set $\Theta$ of unknowns, there is a \emph{most} informative and a \emph{least} informative experiment. Recall the identity function $\id{\Theta}$, $\id{\Theta}(\theta) = \theta$. For any experiment $e$, we have $e \circ \id{\Theta} =  e$. $\id{\Theta}$ provides the decision maker the \emph{exact} value of $\theta$. This experiment has risk, 
$$
\Rbar{L}^\pi(\id{\Theta}) = \EE_{\theta \dist \pi} \min_{a \in A} L(\theta,a).
$$
For any set $X$, define the \emph{terminal} transition $\bullet_X \in\MM(X, \{1\})$ with $\bullet_{X}(x) = 1$ for all $X$. This transition eliminates all information about $X$. For all experiments $e$, $\bullet_{\Theta} = \bullet_{\obs} \circ e$. 
This experiment has risk,
$$
\Rbar{L}^\pi(\bullet_{\Theta}) = \min_{a \in A} \EE_{\theta \dist \pi} L(\theta,a) := \Lbar(\pi). 
$$
By the data processing theorem,
$$
\Rbar{L}^\pi (\id{\Theta}) \leq \Rbar{L}^\pi ( e) \leq \Rbar{L}^\pi (\bullet_{\Theta}).
$$

\section{The General Data Processing Theorem}\label{sec:The General Data Processing Theorem}
For an experiment $e$, define $\MM_{e}(\Theta, A)$ to be the set of all Markov transitions that factor through $e$, transitions of the form $d \circ e$. Key to the functoriality of $\Rbar{L}^\pi$ is the simple realization that if $f : e \rightarrow e'$,
$$
\MM_{e'}(\Theta, A) \subseteq  \MM_{e}(\Theta, A).
$$
This means that the decision maker has more decision rules after performing the experiment $e$ than $e'$, since one could always first inject noise and then proceed as if experiment $e'$ had been performed. This suggests another means to construct quantities that satisfy the data processing theorem.

\begin{theorem}\label{Generalized Information Processing}
	Let $R : \MM(\Theta, A) \rightarrow \RR$ and define,
	$$
	\Rbar{}(e) = \min_{d \in \MM_{e}(\Theta, A)} R(d).
	$$
	If $f :e \rightarrow e'$ then $\Rbar{}(e) \leq \Rbar{}(e')$.
	
\end{theorem}

\begin{proof}
	If $f : e \rightarrow e '$ then $\MM_{e'}(\Theta, A) \subseteq  \MM_{e}(\Theta, A)$. Therefore,
	$$
	\min_{d \in \MM_{e}(\Theta, A)} R(d) \leq \min_{d' \in \MM_{e'}(\Theta, A)} R(d').
	$$
	
\end{proof}
We recover the usual data processing theorem by taking, 
$$
R(d) = \EE_{\theta \dist \pi} \EE_{a\dist d(\theta)}L(\theta, a).
$$ 
The proof of theorem \ref{Generalized Information Processing} makes no explicit reference to expected risks, transitions, or even probability distributions. Then the construction of Theorem \ref{Generalized Information Processing} can be understood as defining a functor from $\MM(\Theta,-)$ to $\left(\RR,\leq\right)$. As we will see, any risk function itself can be seen as a \emph{natural transformation}, and in turn any natural transformation of a particular form yields a general data processing theorem.
\\
\\
Define the constant functor $\RR : \MM(\Theta,-) \rightarrow \mathrm{Set}$, which assigns to each experiment the set of real numbers and to each Markov transition the identity function on the reals. Fix an action set $A$ and consider the following contravariant functor $\mathrm{Decision}_A:\MM(\Theta,-) \rightarrow \mathrm{Set}$, which maps an experiment to its set of decision rules $\MM(\obs, A)$, and for arrows $f : e \rightarrow
e'$, maps a decision rule $d' \in \MM(\obs',A)$ to $d' \circ f \in \MM(\obs, A)$ by pre-processing. \emph{Any} risk function provides a natural transformation,
$$
\risk : \mathrm{Decision}_A \Rightarrow \RR,
$$
with $\risk_{e}(d) = \risk(d\circ e)$. Naturality asserts the diagram below commutes,
	\[
	\xymatrix{
		\RR \ar@{<->}[r]^{\id{\RR}} & \RR \\
		\MM(\obs',A) \ar[u]^{\risk_{e'}} \ar[r]_{\mathrm{Decision}_A(f)} & \MM(\obs, A) \ar[u]_{\risk_e} 
	}
\]
In words, for any decision rule $d'$ and arrow $f : e \rightarrow e'$, 
$$
\risk_{e'}(d') = \risk_e(d' \circ f).
$$
This holds as,
$$
d' \circ e' = d' \circ f \circ e.
$$
\\
Similarly, \emph{any} natural transformation defines a functional risk on $M(\Theta, A)$. As $\id{\Theta}$ is initial in $\MM(\Theta,-)$, naturality asserts the below commutes,
	\[
	\xymatrix{
		\RR \ar@{<->}[r]^{\id{\RR}} & \RR \\
		\MM(\obs,A) \ar[u]^{\risk_{e}} \ar[r]_{\mathrm{Decision}_A(e)} & \MM(\Theta, A) \ar[u]_{\risk_{\id{\Theta}}} 
	}
\]
Meaning, 
$$
\risk_{\id{\Theta}} (e\circ d) = \risk_{e}(d).
$$

\subsection{Presheafs, Risk and Constrained Data Processing Theorems}
Recall the contravariant decision functor,
$$
\mathrm{Decision}_A : \MM(\Theta,-)^{op} \rightarrow \mathrm{Set}
$$
that maps each experiment to the set of decision rules available to the decision maker after performing the experiment. In the literature such a functor is called a presheaf. Currently, this functor provides to the decision maker \emph{all} decision rules $\MM(\obs,A)$. In practice, the decision maker may be constrained to decision rules of a particular form. They could have fairness constraints, unbiased constraints, etc. To model these constraints, we can consider sub-presheafs of the original decision functor, contravariant functors,
$$
D :\MM(\Theta,-)^{op} \rightarrow \mathrm{Set},
$$
with $D(e) \subseteq \MM(\obs,A)$ and for $f :e \rightarrow e'$ a mapping, 
$$
D(f): D(e') \rightarrow D(e)
$$
with $D(f)(d') = d' \circ f$. Modeling of computational constraints will come in future work. Risk remains a natural transformation $\risk : D \Rightarrow \RR$, and our construction from before yields a data processing theorem. We summarize this general result in the following theorem.

\begin{theorem}\label{theorem 5}
    Let $D$ be a presheaf on a category $\mathcal{C}$, and let $\risk : D \Rightarrow \RR$ be a natural transformation. Then
    $$
    \Rbar{}(c) = \min_{d \in D(c)} \risk_c(d),
    $$
    is a functor from $\mathcal{C}$ to $(\RR, \leq)$.
    
\end{theorem}

\begin{proof}
    Let $f : c \rightarrow c'$, and $d' \in D(c')$. Then, by naturality of $\risk$,
    $$
    \min_{d' \in D(c')}\risk_{c'}(d') = \min_{d' \in D(c')} \risk_{c}(D(f)(d')) \geq \min_{d \in D(c)} \risk_{c}(d),
    $$
    where the last inequality is because $D(f)(d') \in D(c)$.
\end{proof}

\subsection{Constructing Presheafs via Admissible Strategies}\label{Presheafs via admissable strategies.}

This construction is based on a single observation: The precomposition leaves the
\emph{strategy} invariant, so any constraint formulated on $\MM(\Theta,A)$ is automatically
stable.

\begin{lemma}[Strategy--level constraints]\label{lem:strategy}
Fix any set $S\subseteq\MM(\Theta,A)$ of admissible strategies and define
\[
   D_{S}(e)\;=\;\bigl\{\,d\in\MM(\obs,A)\;:\;d\circ e\in S\,\bigr\}.
\]
Then $D_{S}$ is a sub-presheaf of the decision presheaf $\mathrm{Decision}_A$ on
$\MM(\Theta,-)$.
\end{lemma}

\begin{proof}
Let $f:e\to e'$ and $d'\in D_{S}(e')$, so $d'\circ e'\in S$. Then
\[
   (d'\circ f)\circ e \;=\; d'\circ(f\circ e)\;=\;d'\circ e'\;\in\;S ,
\]
hence $D(f)(d')=d'\circ f\in D_{S}(e)$. Functoriality is inherited from
$\mathrm{Decision}_A$.
\end{proof}
By theorem \ref{theorem 5} the constrained Bayes risk,
\[
   \Rbar{L,S}^{\pi}(e)=\min_{d\in D_{S}(e)}\riskL{L}^\pi(e,d),
\]
is then a functor $\MM(\Theta,-)\to(\RR,\leq)$.

\subsubsection{From General Presheaves to Strategy-Level Constraints}

As $\id{\Theta}$ is initial in $\MM(\Theta,-)$, any experiment $e$ is naturally an arrow from $\id{\Theta} \rightarrow e$. Take $S = D(\id{\Theta})$. Applying $D(e): D(e) \rightarrow D(\id{\Theta})$, with $d \mapsto d \circ e \in D(\id{\Theta})= S$, meaning that $d \in D(e)$ implies $d \in D_S(e)$. Therefore all sub-presheafs of $\mathrm{Decision}_A$ are included in a canonical strategy-level constraint.

\subsection{Bifunctors, Post Processing and Constrained Deficiency}

Thus far we have considered the action set to be \emph{static}. Here we alow the actions themselves to change via postprocessing. Let, 
$$
D : \MM(\Theta,-)^{op} \times \underbrace{\mathrm{StochRel}}_{actions} \rightarrow \mathrm{Set},
$$
be a \emph{bifunctor}, the canonical example of which is,
$$
D(e,A) = \{d \in \MM(\obs,A) :d\circ e \in S_A\},
$$
with $S_A$ an action set dependent subset of $\MM(\Theta, A)$. Such functors reflect constraints that are closed under post-processing. Let $h : A \rightarrow A'$. We have a map,
\begin{align*}
    D(e,A) &\rightarrow D(e,A') \\
    d &\mapsto h\circ d \in S_{A'}.
\end{align*}
We assume further that $D(e,A)$ is \emph{convex}. This is needed for the proof of theorem \ref{Constrained Randomization Theorem} and is completely natural if one allows randomized decision rules. Fairness and privacy constraints fit the bill, unbiasedness does not as it is not closed under post processing.

\subsubsection{Constrained Deficiency}

Recall the notion of \emph{deficiency} \citep{Le1964,Torgersen1991,vanrooyen2025experiments},
$$
\xi^{\pi}(e, e') := \min_{f \in \MM(\obs, \obs')} \EE_{\theta \dist \pi} V(f\circ e(\theta), e'(\theta)),
$$
where $V$ is the \emph{variational divergence}. Deficiency measures how close $e'$ is to being $e$ with added noise. Deficiency quantifies the maximum difference in risks between two experiments as we allow the loss to vary,
$$
\Rbar{L}^\pi(e) \leq \Rbar{L}^\pi(e') +\norm{L}_\infty \xi^{\pi}(e, e').
$$
Crucially, deficiency is predicated on access to \emph{all} decision rules. What we want is a bound,
$$
\Rbar{L,D}^\pi(e) \leq \Rbar{L,D}^\pi(e') + \norm{L}_\infty \xi^{\pi}_{D}(e, e'),
$$
that factors in the need to work with restricted decision rules. Here we present such a bound. Our plan of attack will be as follows:

\begin{enumerate}
    \item We construct a set $H(e)$ for each experiment with,
    $$
    H(e) = \{h \in \MM(\obs, \tilde{\obs}) : h\circ e \in S_{\tilde{Z}} \}
    $$
    in words, we are looking at transformations $h$ that would render the post-processed version of $e$ compliant with constraints. Crucially $\tilde{\obs}$ is not fixed apriori, we are considering all morphisms coming out of $\obs$.

    \item $H(e)$ is non-empty as $D(e,A)\subseteq H(e)$.

    \item We notice that for \emph{any} downstream processing $d \in \MM(\tilde{\obs}, A)$, $d\circ h \in S_A$, for $h \in H(e)$.

    \item This puts us back in the unconstrained space, where we can compare experiments via the deficiency. 
    
\end{enumerate}

We will see that, 
$$
\xi^{\pi}_{D}(e, e') = \max_{h' \in H(e')} \min_{h \in H(e)} \xi^{\pi}(h \circ e, h'\circ e'),
$$
with $h$ and $h'$ the decompositions of $e$ and $e'$ accordingly. 

\begin{theorem}\label{Constrained Randomization Theorem}
Let $e$ and $e'$ be experiments with, 
$$
D(e,A) = \{d \in \MM(\obs,A) :d\circ e \in S_A\},
$$
a bifunctor, for some strategy set map $S$. Then,
$$
\Rbar{L,D}^\pi(e) \leq  \Rbar{L,D}^\pi(e') + \epsilon \norm{L}_\infty,
$$
for all loss functions and action sets if and only if, 
$$
\xi^\pi_D(e,e') \leq \epsilon.
$$

\end{theorem}

\begin{proof}
    We closely follow the proof of the randomization theorem given in \citep{vanrooyen2025experiments} with some modifications. First suppose,
    $$
    \xi^\pi_D(e,e') \leq \epsilon.
    $$
    Then for \emph{any} decision rule $\mathbb{d}' \in D(e',A)$, we can write $\mathbb{d}' = d' \circ h'$. We are guaranteed a $f$ and a $h$,
    $$
    \EE_{\theta \dist \pi} V(f\circ h \circ e(\theta) ,h'\circ e'(\theta))\leq \epsilon.
    $$
    Then,
    \begin{align*}
            \riskL{L}^\pi(e, d' \circ f \circ h) - \riskL{L}^\pi(e', d' \circ h') &\leq \norm{L}_\infty \EE_{\theta \dist \pi} V(d' \circ f \circ h \circ e(\theta), d'\circ h' \circ e'(\theta)) \\
            &\leq \norm{L}_\infty \EE_{\theta \dist \pi} V(f \circ h \circ e(\theta),h' \circ e'(\theta))  \\
            &\leq \norm{L}_\infty \epsilon 
    \end{align*}
\\
\\
The proof is completed by taking minima over $d'$. For the converse, consider the function,

$$
\phi(L,d) = \riskL{L}^\pi(e, d) - \riskL{L}^\pi(e', d') - \epsilon \norm{L}_\infty.
$$
with $d' \circ e' \in S_A$. Note that $\phi$ is affine in $d$ and concave in $L$. By the conditions in the theorem,
$$
\sup_{L} \min_{d\in D(e,A)} \phi(L,d) \leq 0.
$$
By the minimax theorem \citep{Komiya1988} or strong convex duality \citep{Lucchetti2006}, there exists a saddle point $(L^*, d^*)$ with, 
$$
\phi (L^*, d^*) = \min_{d \in D(e,A)} \sup_{L} \phi(L,d) = \sup_{L} \min_{d \in D(e,A)} \phi(L,d) \leq 0  .
$$
This implies,
$$
\riskL{L}^\pi(e, d^*) \leq \riskL{L}^\pi(e', d') + \epsilon \norm{L}_\infty ,\ \forall L.
$$
This means $\EE_{\theta \dist \pi} V(d^* \circ e(\theta) , d' \circ e'(\theta) ) \leq \epsilon$. To complete the proof, let $A = \tilde{\obs}$ and $d' = h'$. Then $d^* = d \circ h$ for some $h$ and,
$$
\EE_{\theta \dist \pi} V(d \circ h \circ e(\theta) , h' \circ e'(\theta) ) \leq \epsilon.
$$
Therefore $f$ is given by $d$.

\end{proof}
\\
Setting $\epsilon = 0$ in the above yields the following corollary, a constrained Blackwell-Sherman-Stein theorem.

\begin{corollary}[Constrained BSS]
$$
\Rbar{L,D}^\pi(e) \leq  \Rbar{L,D}^\pi(e')
$$
for all loss functions if and only if,
$$
\xi^\pi_D(e,e') = 0.
$$

\end{corollary}

\subsubsection{Minimal Transformations in H}

Intuitively, $h \in H(e)$ throws away information in $e$ that is not relevant to enforcing the constraints. One need not consider \emph{all} $H$ in making decisions, rather only those with that are \emph{minimal}. $H(e)$ is naturally a preorder, with $h_1 \leq h_2$ if there exists an arrow $f : h_1 \rightarrow h_2$. $h \in H(e)$ is \emph{minimal} if it is less than every $h'$ it can be compared with. Denote by $\underline{H}(e)$ the minimal elements of $H(e)$. We have the following theorem,

\begin{theorem}
    $$
\xi^\pi_D(e,e') = \max_{h' \in \underline{H}(e')} \min_{h \in \underline{H}(e)} \xi^{\pi}(h \circ e, h'\circ e').
$$
\end{theorem}

\begin{proof}
    We need to show that the minima of,
    $$
     \max_{h' \in H(e')} \min_{h \in H(e)} \xi^{\pi}(h \circ e, h'\circ e'),
    $$
    occur when $h$ and $h'$ are minimal. Define
    $$
    \Phi(h') = \min_{h \in H(e)} \xi^{\pi}(h \circ e, h'\circ e').
    $$
    suppose that $h_1 \leq h_2$, ie $f: h_1 \rightarrow h_2$. We have, 
    \begin{align*}
            \xi^{\pi}(h_2 \circ e, h'\circ e') &= \min_{k} \EE_{\theta\dist \pi} V(k \circ h_2\circ e(\theta), h' \circ e'(\theta)) \\
            &= \min_{k} \EE_{\theta\dist \pi} V(k \circ f \circ h_1\circ e(\theta), h' \circ e'(\theta)) \\
            &\geq \min_g \EE_{\theta\dist \pi} V(g \circ h_1\circ e(\theta), h' \circ e'(\theta)) \\
            &= \xi^{\pi}(h_1 \circ e, h'\circ e').
    \end{align*}
    Therefore the inner minima is obtained at a minimal point. Now suppose $h'_1 \leq h'_2$. Then,
    Fix $h$ to $h^*$, obtaining the inner minima at $h'_1$. Then,
    \begin{align*}
        \Phi(h'_2) &\leq \min_k\EE_{\theta\dist \pi} V(k \circ h^* \circ e(\theta), h_2' \circ e'(\theta)) \\
        &= \min_k\EE_{\theta\dist \pi} V(k \circ h^* \circ e(\theta), f \circ h_1' \circ e'(\theta)) \\
        &\leq \min_g\EE_{\theta\dist \pi} V(f \circ g \circ h^* \circ e(\theta), f \circ h_1' \circ e'(\theta)) \\
        &\leq \min_g\EE_{\theta\dist \pi} V(g \circ h^* \circ e(\theta), h_1' \circ e'(\theta)) \\
        &= \Phi(h'_1).
    \end{align*}
    This in turn means the outer maximiser is obtained at a minimal point.    
\end{proof}
\\
Returning to the unconstrained case, we have $S_\obs = \MM(\Theta,\obs)$, meaning $\id{\obs} \in H(e)$.  It is clear that $\id{\obs}$ is also minimal meaning in this case,
$$
\xi^\pi_D(e, e') = \xi^\pi(e,e').
$$

\section{Some Tangible Constructions in Statistics}

\subsection{Example: Binary Experiments and the Variational Divergence}

Let $\Theta = A =\{-1,1\}$ with $L(\theta,a) = \pred{\theta \neq a}$. In words, there are two unknowns, and all the loss function cares about is that we accurately identify which of the two it is. Let $e$ be an experiment and $\pi$ the uniform prior on $\Theta$. Then,
\begin{align*}
	\Rbar{L}^\pi(e) &= \min_{f\in \{-1,1\}^\obs} \EE_{\theta \dist \pi} \EE_{x\dist e(\theta)}L(\theta,f(x)) \\
	&= \min_{f\in \{-1,1\}^\obs} \frac{1}{2} \left(\EE_{e(-1)}\pred{1 = f(Z)} + \EE_{e(1)} \pred{-1 = f(Z)}\right)
\end{align*}
Let $C = \{z \in \obs : f(z) = 1\}$, then
\begin{align*}
	\Rbar{L}^\pi(e) &= \frac{1}{2} \min_{C} \left(e_{-1}(\ind_C) + e_{1}(\ind_{C^c}) \right) \\
	& = \frac{1}{2} \min_{C} \left(e_{-1}(\ind_C) + 1 - e_{1}(\ind_{C}) \right).
\end{align*}
This yields,
\begin{align*}
	\Rbar{L}^\pi(\bullet_{\Theta}) - \Rbar{L}^\pi(e) &= \frac{1}{2} - \frac{1}{2} \min_{C} \left(e_{-1}(\ind_C) + 1 - e_{1}(\ind_{C}) \right) \\
	&= \frac{1}{2} \max_{C} \left(e_{1}(\ind_C) - e_{-1}(\ind_{C}) \right) \\
	&= V(e_{-1},e_1),
\end{align*}
which is the total variational distance between $e(-1)$ and $e(1)$, a commonly used metric between probability distributions. The data processing theorem recovers the well-known fact that,
$$
V(P,Q) \geq V(f(P),f(Q)),
$$
for all distributions $P,Q \in \PP(\obs)$ and all Markov kernels $f \in \MM(\obs,\obs')$.

\subsection{Example: Connections to \texorpdfstring{$\phi$}{phi}-divergences and multi-\texorpdfstring{$\phi$}{phi}-divergences.}

Let $\phi : [0,\infty) \rightarrow \RR$ be a convex function with $\phi(1)=0$. Then the $\phi$-divergence between the distributions $P,Q \in \PP(\obs)$ is given by,
$$
D_\phi(P,Q) = \EE_{P}\phi\left(\frac{dQ}{dP}\right),
$$
if $Q \ll P$ and $\infty$ otherwise. As we have seen, for binary experiments, for a particular loss and prior, the gap,
$$
\Rbar{L}^\pi(\bullet_{\Theta}) - \Rbar{L}^\pi(e) = V(e_{-1},e_{1}).
$$
The variational divergence is an example of a $\phi$-divergence with $\phi(x) = |x - 1|$. This construction works more broadly, for binary experiments the gap,
$$
\Rbar{L}^\pi(\bullet_{\Theta}) - \Rbar{L}^\pi(e) = D_\phi(e_{-1},e_{1}),
$$
for some $\phi$. Similarly, for every $\phi$-divergence, there is a loss and prior that makes the above hold. We refer the reader to \citep{Reid2009b} for the details of this correspondence. Our general data processing theorem then recovers the fact,
$$
D_\phi(P,Q) \geq D_\phi (f(P),f(Q)).
$$
Similarly, if $|\Theta| > 2$ we can invoke a similar correspondence between,
$$
\Rbar{L}^\pi(\bullet_{\Theta}) - \Rbar{L}^\pi(e)
$$
and objects called multi-$\phi$-divergences. We refer the reader to \citep{Garca-Garca}. Our general data processing theorem once again recovers the data processing theorem for multi-$\phi$-divergences.

\subsection{Example: Unbiasedness and Data Processing}
Let $g : \Theta \rightarrow A$, where $A$ is a convex subset of $\RR^n$. A strategy $s \in \MM(\Theta, A)$, is unbiased for $g$ if,
\[
\EE_{a\dist s(\theta)} a = g(\theta).
\]
Intuitively these estimates get $g$ correct on average. Usually in proving data processing inequalities we work with \emph{all} possible decision rules, here we recover a data processing theorem when we restrict to those rules that are unbiased. Let $S \subseteq \MM(\Theta,A)$ be the set of all unbiased strategies, then the construction of section \ref{Presheafs via admissable strategies.} yields a presheaf $D_S$, with $D_S(e)$ comprising all decision rules that are unbiased, $d \circ e \in S$. Using the standard Bayes risk for an experiment yields functoriality of,
\[
\Rbar{L,S}^{\pi}(e)=\min_{d\in D_{S}(e)}\riskL{L}^\pi(e,d),
\]
ie if we restrict to unbiased strategies, then we still yield a data processing inequality.

\subsection{Example: Fairness and Data Processing}
Suppose that our set of unknowns factors as $\Theta = \Theta_0 \times Y$, where $Y$ is some protected attribute that we do not wish to be able to predict from the output of our overall strategy. For example $\Theta_0 = \{\text{Default},\text{Pay}\}$, could be the unknown future state of some loan and $Y$ the gender of the applicant. If we can \emph{infer $Y$} from the output of the strategy, then we are not using gender information fairly, our decision is biased by this side information. In fairness aware learning, we seek to minimize these issues, to find decision rules that are themselves fair over certain protected attributes.

\subsubsection{Notions of Fairness}
There are many notions of fairness in the literature, here we focus on \emph{demographic parity}. Fix a prior $\pi \in \PP(\Theta)$. A strategy $s \in \MM(\Theta, A)$ induces a joint distribution over $\Theta \times A$ in the usual way. A strategy is \emph{fair} if,
\[
A \perp Y,
\]
the actions chosen by the algorithm are independent of the protected attribute. This defines a strategy set,
\[
S := \left\{ s \in \MM(\Theta,A) : A \perp Y \text{ under } \pi \right\}.
\]
Fixing a loss $L$ yields functoriality of,
\[
\Rbar{L,S}^{\pi}(e)=\min_{d\in D_{S}(e)}\riskL{L}^\pi(e,d),
\]
in words, noise degrades performance even when restricted to fair rules.

\subsubsection{A Worked Example: Group-Targeted Noise}
\label{sec:worked-example}

The constrained inequalities of the preceding sections are produced by the
same one-line containment argument as the classical data-processing theorem,
and one may reasonably ask whether they have any content the unconstrained
theory lacks. The following example shows that they do: a constrained risk
functional can separate, at a \emph{single} loss and prior, a pair of
experiments that the corresponding unconstrained Bayes risk identifies, and
in doing so furnishes a one-line certificate of Blackwell-incomparability.
\\
\\
The constraint used is \emph{equalized odds} rather than demographic parity.
Equalized odds is a strategy-level constraint in the sense of
Lemma~\ref{lem:strategy}: writing $\Theta = \Theta_0 \times Y$
as before, a strategy $s \in M(\Theta, A)$ satisfies equalized odds if,
under the joint distribution induced by the prior $\pi$,
\[
A \perp Y \mid \Theta_0 ,
\]
that is, the action is conditionally independent of the protected attribute
given the task label. This is a property of $s$ and $\pi$ alone, so
$S_{\mathrm{EO}} := \{ s \in M(\Theta, A) : A \perp Y \mid \Theta_0
\text{ under } \pi \}$
defines a sub-presheaf $D_{S_{\mathrm{EO}}}$ by
Lemma~\ref{lem:strategy}, and
Theorem~\ref{theorem 5} applies verbatim.

\paragraph{Setup.}
Let $\Theta_0 = Y = \{0,1\}$, let $\pi$ be uniform on
$\Theta = \Theta_0 \times Y$ (so $\Theta_0 \perp Y$ under the prior, with
equal base rates across groups), let $A = \{0,1\}$, and let
$L(\theta, a) =  \pred{a \neq \theta_0}$ be the $0$--$1$ loss
on the task label. Fix $\varepsilon \in (0, \tfrac{1}{2})$. Both experiments
have observation space $Z = \{0,1\}$:
\begin{itemize}
  \item \textbf{Group-blind noise.} The experiment
  $e_1 \in M(\Theta, Z)$ reports $\theta_0$ through a binary symmetric
  channel of crossover probability $\varepsilon$, independently of $y$:
  \[
  \mathbb{P}(Z = \theta_0 \mid \theta_0, y) = 1 - \varepsilon .
  \]
  \item \textbf{Group-targeted noise.} The experiment
  $e_2 \in M(\Theta, Z)$ reports $\theta_0$ perfectly on one group and
  noisily on the other:
  \[
  \mathbb{P}(Z = \theta_0 \mid \theta_0, y = 0) = 1, \qquad
  \mathbb{P}(Z = \theta_0 \mid \theta_0, y = 1) = 1 - 2\varepsilon .
  \]
\end{itemize}
The two experiments inject the same total amount of noise into the task
label; they differ only in where that noise lives.

\begin{proposition}[Constrained risk separates; unconstrained risk does not]
\label{prop:separation}
With the setup above,
\[
\mathrm{Risk}^{\pi}_{L}(e_1)
  \;=\; \mathrm{Risk}^{\pi}_{L}(e_2)
  \;=\; \varepsilon ,
\]
while
\[
\mathrm{Risk}^{\pi}_{L, S_{\mathrm{EO}}}(e_1) \;=\; \varepsilon ,
\qquad
\mathrm{Risk}^{\pi}_{L, S_{\mathrm{EO}}}(e_2)
  \;=\; \tfrac{1}{2}
  \;=\; \mathrm{Risk}^{\pi}_{L}(\bullet_\Theta) .
\]
\end{proposition}

\begin{proof}
\emph{Unconstrained risks.}
For $e_1$ the posterior over $\Theta_0$ given $Z = z$ places mass
$1 - \varepsilon$ on $z$, so the rule $a = z$ is Bayes with risk
$\varepsilon$. For $e_2$, averaging over the uniform $Y$,
\[
\mathbb{P}(Z = z \mid \Theta_0 = z)
  = \tfrac{1}{2} \cdot 1 + \tfrac{1}{2}(1 - 2\varepsilon)
  = 1 - \varepsilon ,
\]
so the posterior is identical to that of $e_1$, the rule $a = z$ is again
Bayes, and its risk is
$\tfrac{1}{2}\cdot 0 + \tfrac{1}{2}\cdot 2\varepsilon = \varepsilon$.
\\
\\
\emph{Constrained risk of $e_1$.}
The Bayes rule $a = z$ already lies in $D_{S_{\mathrm{EO}}}(e_1)$:
conditional on $\Theta_0 = \theta_0$, the action is
$\theta_0$ flipped by noise drawn independently of $Y$, so
$A \perp Y \mid \Theta_0$. Since the constrained minimum dominates the
unconstrained one and is here achieved,
$\mathrm{Risk}^{\pi}_{L, S_{\mathrm{EO}}}(e_1) = \varepsilon$.
\\
\\
\emph{Constrained risk of $e_2$.}
Parametrize a decision rule $d \in M(Z, A)$ by
$p_z = \mathbb{P}(A = 1 \mid Z = z)$. Conditional on $\Theta_0 = 0$:
if $Y = 0$ then $Z = 0$ surely, so
$\mathbb{P}(A = 1 \mid \Theta_0 = 0, Y = 0) = p_0$;
if $Y = 1$ then
$\mathbb{P}(A = 1 \mid \Theta_0 = 0, Y = 1)
  = (1 - 2\varepsilon)\,p_0 + 2\varepsilon\, p_1$.
Equalized odds requires these to coincide, i.e.
$2\varepsilon\,(p_1 - p_0) = 0$, hence $p_1 = p_0$ (the case
$\Theta_0 = 1$ yields the same condition by symmetry). Thus
$D_{S_{\mathrm{EO}}}(e_2)$ consists exactly of the constant randomized
rules $p_0 = p_1 = p$, every one of which has risk
\[
\mathbb{P}(A \neq \Theta_0)
  = \tfrac{1}{2}\, p + \tfrac{1}{2}\,(1 - p)
  = \tfrac{1}{2} ,
\]
independent of $p$. Hence
$\mathrm{Risk}^{\pi}_{L, S_{\mathrm{EO}}}(e_2) = \tfrac{1}{2}$, the risk of
the trivial experiment.
\end{proof}
\\
\\
Under the equalized-odds constraint, the fibre over $e_2$ collapses to the
rules that ignore the data entirely: group-targeted noise, although
invisible to the task-averaged Bayes risk, is catastrophic for a decision
maker required to act fairly.

\begin{corollary}[An incomparability certificate]
\label{cor:certificate}
For every $\varepsilon \in (0, \tfrac{1}{2})$, the experiments $e_1$ and
$e_2$ are Blackwell-incomparable.
\end{corollary}

\begin{proof}
If $e_1$ were a garbling of $e_2$, functoriality of the constrained risk
(Theorem~\ref{theorem 5} with Lemma~\ref{lem:strategy})
would give
\[
\tfrac{1}{2}
  = \mathrm{Risk}^{\pi}_{L, S_{\mathrm{EO}}}(e_2)
  \leq \mathrm{Risk}^{\pi}_{L, S_{\mathrm{EO}}}(e_1)
  = \varepsilon ,
\]
a contradiction. Conversely, $e_1(\theta_0, y)$ does not depend on $y$,
so neither does any garbling $f \circ e_1$; since $e_2(\theta_0, \cdot)$
genuinely depends on $y$ for $\varepsilon > 0$, $e_2$ is not a garbling of
$e_1$.
\end{proof}
\\
\\
The first half of the corollary is the point. By the Blackwell--Sherman--Stein
theorem, certifying non-domination through unconstrained risks requires
exhibiting \emph{some} decision problem on which the risks reverse, and the
natural problem here is silent: Proposition~\ref{prop:separation} shows the
unconstrained risks at $(L, \pi)$ agree. A single constrained functional,
at that same loss and prior, settles the question. (A direct verification
is also instructive: any factorization $e_1 = f \circ e_2$ is pinned down
on the $y = 0$ fibre, where $e_2$ is deterministic, to
$f = \mathrm{BSC}_\varepsilon$; the $y = 1$ fibre then demands
$4\varepsilon^2 = 2\varepsilon$, i.e.\ $\varepsilon = \tfrac{1}{2}$.)

\subsubsection{Approximate Notions of Fairness}
Our current notion of fairness is a hard constraint, following \citep{pmlr-v81-menon18a} we consider approximate notions of fairness. Let $\mathrm{F} : \MM(\Theta, A) \rightarrow \RR$ be a function that says how unfair a strategy is.
Let,
\[
S_\alpha := \left\{s : F(s) \leq \alpha \right\},
\]
those strategies that have unfairness less than $\alpha$. If $\alpha = 0$ we return to the case of the previous section, if $\alpha = \infty$ we are in the unrestricted regime. Then,
\[
\Rbar{L,S_{\alpha}}^{\pi}(e)=\min_{d\in D_{S_\alpha}(e)}\riskL{L}^\pi(e,d),
\]
is a functor to $(\overline{\mathbb{R}},\leq)$. Therefore, even if we constrain unfairness to be less than $\alpha$, risk increases under noise.

\section{Conclusion}
The informativeness of an experiment is not a property it holds in isolation but a relation: one experiment dominates another exactly when it offers the decision maker a larger set of decision rules. Cast this way, the data-processing inequality is the functoriality of fibrewise minimisation over the decision presheaf, and the classical inequalities for total variation and f-divergences fall out as instances of a proof that makes no reference to probability.
\\
\\
The point of the reformulation is that it continues to hold over sub-presheaves. Constraints imposed on induced strategies — fairness, unbiasedness — are automatically of this kind, so a decision maker required to act fairly still enjoys a data-processing inequality. For constraints additionally closed under post-processing, the constrained deficiency and the constrained randomization theorem supply the quantitative form, and the equalized-odds example shows that the constrained theory separates experiments the unconstrained theory identifies.
\\
\\
Several things remain open. The constrained deficiency is defined as an optimisation over compliant coarsenings, reduced by Theorem 9 to the minimal ones; we do not know how to characterise those minimal elements for the standard fairness notions, and without such a characterisation the quantity is hard to compute even in small examples. Whether the constrained deficiency metrises a useful topology on experiments, as Le Cam distance does in the unconstrained case, is likewise unsettled. Theorem 7 requires convexity of the admissible fibres, which excludes unbiasedness; whether a constrained randomization theorem holds for non-convex admissible classes we leave open. Finally, the framework is basis-free and should survive the passage to infinite observation spaces, though the measure-theoretic details deserve care.

\newpage

\bibliographystyle{plainnat}
\bibliography{./References}

\end{document}